\documentclass[10pt, conference,letterpaper]{IEEEtran}
\makeatletter
\def\ps@headings{
\def\@oddhead{\mbox{}\scriptsize\rightmark \hfil 
}
\def\@evenhead{\scriptsize
\hfil \leftmark\mbox{}}
\def\@oddfoot{}
\def\@evenfoot{}}
\usepackage{booktabs}
\makeatother 

\usepackage{colortbl}
\usepackage{xcolor}
\usepackage{array}

\usepackage{graphicx}
\usepackage{epstopdf}
\usepackage{cite} 
\usepackage{times}
\usepackage{dsfont} 
\usepackage{cite}
\usepackage{times}
\usepackage{pifont}
\usepackage{multirow}   
\usepackage{tikz}
\usetikzlibrary{shapes.geometric,calc}
   
\usepackage{amsthm}
\usepackage{graphicx}
\usepackage{subfigure}
\usepackage{color}
\usepackage{ifpdf}
\usepackage{epsfig}
\usepackage{latexsym}
\usepackage{amsfonts}
\usepackage{amssymb}
\usepackage{paralist}
\usepackage{comment}
\usepackage{xspace}
\usepackage{mathrsfs}
\usepackage{amssymb}
\usepackage{setspace}
\usepackage{color}
\usepackage[small]{caption}

\usepackage{subeqnarray}
\usepackage{amssymb}
\usepackage{url,epsfig,array}
\usepackage{leftidx}
\usepackage{amsmath}
\usepackage[T1]{fontenc}
\usepackage{aecompl}

\usepackage{calligra}
\usepackage{algorithm}
\usepackage{algorithmicx}
\usepackage{algpseudocode} 
\usepackage{amsmath}

\def\ie{\textit{i.e.}\xspace}
\def\etal{\textit{et al.}\xspace}
\def\etc{\textit{etc.}\xspace}

\def\eg{\textit{e.g.}\xspace}

\makeatletter
\renewcommand{\maketag@@@}[1]{\hbox{\m@th\normalsize\normalfont#1}}%
\makeatother

\begin{document}
  
\title{\LARGE Fair Differentiable Neural Network Architecture Search for Long-Tailed Data with Self-Supervised Learning}

\author{
\IEEEauthorblockN{Jiaming Yan}
\IEEEauthorblockA{\textit{School of Computer Science and Technology}
\\ \textit{University of Science and Technology of China}}
}

\maketitle

\begin{abstract}
Recent advancements in artificial intelligence (AI) have positioned deep learning (DL) as a pivotal technology in fields like computer vision, data mining, and natural language processing. 
A critical factor in DL performance is the selection of neural network architecture. 
Traditional predefined architectures often fail to adapt to different data distributions, making it challenging to achieve optimal performance. 
Neural architecture search (NAS) offers a solution by automatically designing architectures tailored to specific datasets. 
However, the effectiveness of NAS diminishes on long-tailed datasets, where a few classes have abundant samples, and many have few, leading to biased models.
In this paper, we explore to improve the searching and training performance of NAS on long-tailed datasets. 
Specifically, we first discuss the related works about NAS and the deep learning method for long-tailed datasets.
Then, we focus on an existing work, called SSF-NAS, which integrates the self-supervised learning and fair differentiable NAS to making NAS achieve better performance on long-tailed datasets.
An detailed description about the fundamental techniques for SSF-NAS is provided in this paper, including DARTS, FairDARTS, and Barlow Twins.
Finally, we conducted a series of experiments on the CIFAR10-LT dataset for performance evaluation, where the results are align with our expectation.
\end{abstract}
 

\section{Introduction}\label{sec:introduction}
With the rapid development of artificial intelligence (AI) in recent years, deep learning (DL) has drawn an unprecedented level of attention with groundbreaking successes in numerous fields, including computer vision \cite{voulodimos2018deep}, data mining \cite{nguyen2019machine}, and natural language processing \cite{otter2020survey}.
For the deep neural networks (DNNs) training, the selection of neural network architecture has a significant impact on training performance (\eg, accuracy).
For example, a too-simple architecture may lack sufficient representational capacity, leading to poor training performance.
On the contrary, a too-complex architecture will bring unnecessary computational overhead, as well as the risk of overfitting.
In a word, it is important for the DL application to select a suitable architecture.

Given a target dataset, most of the existing works still adopt the predefined neural network architectures, such as ResNet \cite{he2016deep}, MobileNet \cite{sandler2018mobilenetv2}, and \etc.
However, this will encounter two non-negligible problems.
Firstly, the adopted architecture may not be the optimal for the target dataset.
As studied in the previous research \cite{luo2021architecture}, the architecture that achieves the best performance on a specific data distribution may not generalize well to others.
We already know that an inappropriate architecture can reduce the training performance significantly.
Secondly, it is very difficult for model developers to design the optimal architecture over the target dataset.
There are a lot of trials and errors in the design process, which demands extensive human expertise, effort, and time cost.
Thus, manually designing a suitable architecture for every learning task is unrealistic.
In order to tackle these two key challenges, the neural architecture search (NAS) technique is proposed \cite{zoph2016neural}, which can automatically generate the optimal architecture on the target dataset.
Recently, the field of NAS has achieved several revolutionary breakthroughs, and received significant interest from both research and industrial communities.
At present, the automatically designed architectures drived by NAS have shown competitive performance compared to the predefined architectures in many areas, including image classification, object detection, and semantic segmentation \cite{elsken2019neural}.

\begin{figure}[t]\centering
    \includegraphics[width=0.45\textwidth]{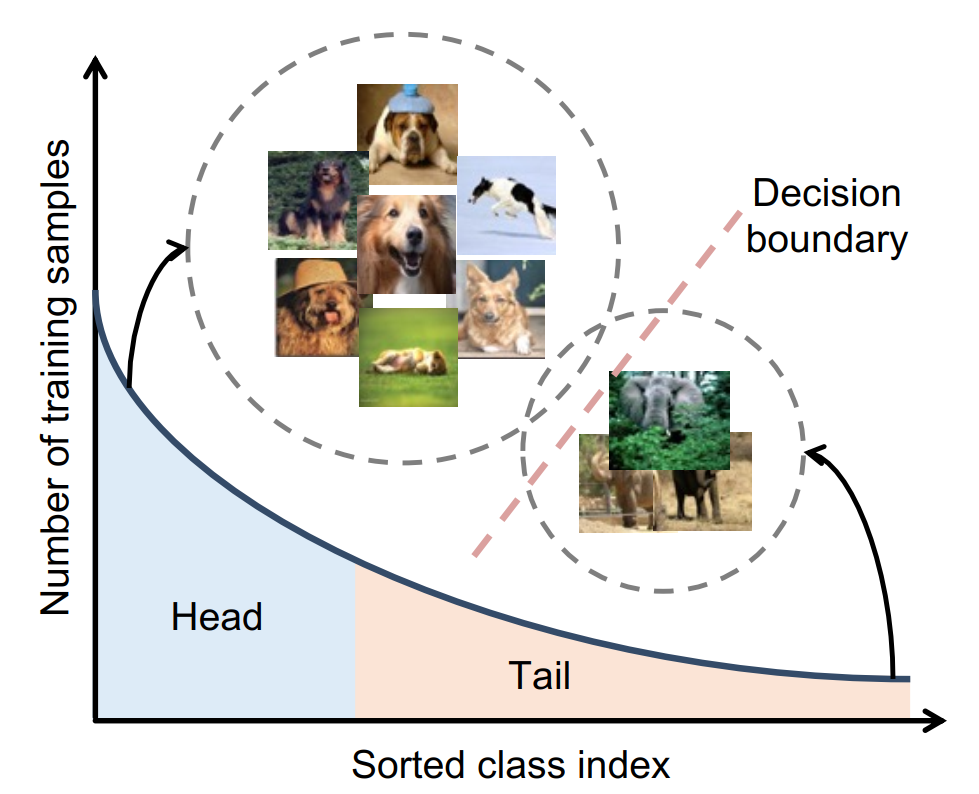}
    \caption{The Label Distribution of a Long-Tailed Dataset \cite{zhang2023deep}.} \label{fig:long_tailed_data}
 \end{figure}

Although NAS has brought great convenience and performance improvement to the DL application, it may only be effective on the datasets that are balanced across classes.
However, in practical, the target datasets usually exhibit a long-tailed class distribution, in which a few classes contain a large number of sample, while the majority have only a few samples.
For example, as shown in Fig. \ref{fig:long_tailed_data}, the number of dog images is much larger than that of elephant images.
When searching and training over such imbalanced datasets, the model can be easily biased towards head classes with massive samples, leading to low accuracy on tail classes that have scarce samples.
For instance, Duggal \etal \cite{duggal2022imb} trained the same architecture over the datasets with different distributions.
Upon altering the data distribution from balance to long-tailed, the test accuracy of this architecture decreased significantly, ranging from a 17.3\% to 36.7\% reduction.

Therefore, in this paper, our primary objective is to improve the searching and training performance of NAS on the datasets with long-tailed distribution.
Specifically, we learn the scheme proposed in \cite{timofeev2021self}, which integrates the fair differentiable NAS technique and the self-supervised learning method.
Then, we implement this scheme over the imbalanced CIFAR dataset \cite{krizhevsky2009learning} to evaluate its performance.
The rest of this paper is organized as follows.
Section \ref{sec:related_work} disscuss the previous works about neural architecture search and deep long-tailed learning.
In Section \ref{sec:methodology}, we introduce some preliminaries about the related techniques, including DARTS \cite{zoph2016neural}, FairDARTS \cite{chu2020fair}, and Barlow Twins \cite{zbontar2021barlow}, and then presents a detailed description of the scheme in \cite{timofeev2021self}.
Subsequently, the experimental setup and evaluation results are shown in Section \ref{sec:experiments}.
Finally, we conclude this paper in Section \ref{sec:conclusion}.

\section{Related Work} \label{sec:related_work}

\subsection{\textbf{Neural Architecture Search}} \label{sec:related_work_nas}
\begin{figure}[h]\centering
    \includegraphics[width=0.5\textwidth]{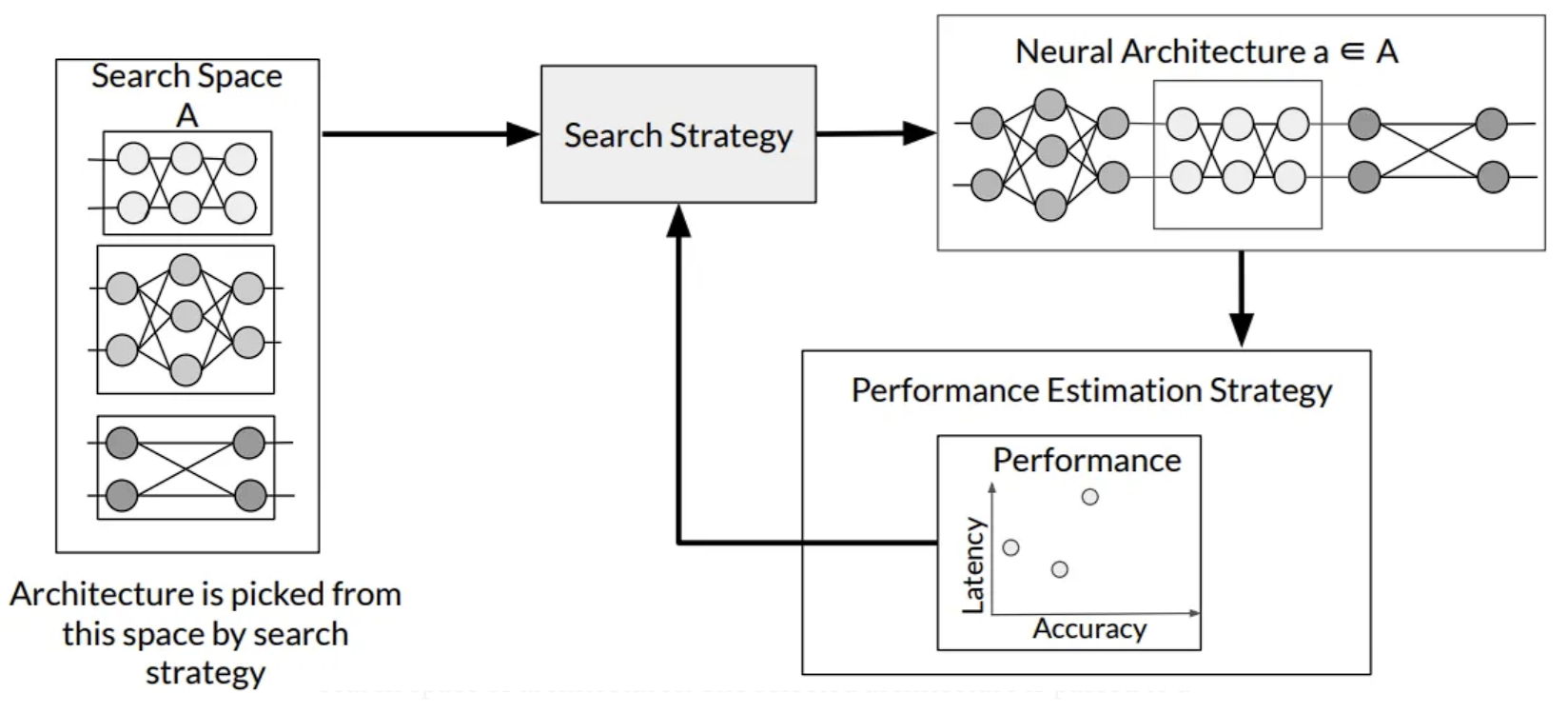}
    \caption{The Process of NAS.} \label{fig:nas_process}
 \end{figure}

Generally speaking, the process of NAS is composed of three main steps, which are illustrated in Fig. \ref{fig:nas_process}\footnote{https://medium.com/@ashwinkumarjs/neural-architecture-search-df415d451fed}.
At the beginning, we first define a search space that contains all the candidate architectures, and initialize the search strategy $\mathcal{S}^{1}$ as well as the performance estimation strategy $\mathcal{E}$.
Then, the NAS process is conducted iteratively.
In each iteration $t$:

\begin{itemize}
    \item \textbf{Step 1.} The search strategy $\mathcal{S}^{t}$ first picks an architecture $a^{t}$ from the predefined search space $A$.
    \item \textbf{Step 2.} The selected architecture $a$ is evaluated by the performance estimation strategy $\mathcal{E}$.
    \item \textbf{Step 3.} The performance of architecture $a^{t}$ (\eg, inference latency, test accuracy) is utilized to update the search strategy, \ie, $\mathcal{S}^{t} \rightarrow \mathcal{S}^{t+1}$.
\end{itemize}

By conducting these steps repeatedly, the search strategy can be optimized continuously.
As a result, the optimal architecture with the best evaluation performance will be selected.

The existing NAS methods can be roughly divided into the following three categories according to their different search strategies: reinforcement learning (RL), evolutionary algorithm (EA), and gradient-based (GD) \cite{zhu2021federated}. 
Zoph \etal \cite{pham2018efficient} first adopt the RL to train a recurrent neural network (RNN) model that generates architectures, which is a pioneering work in the field of NAS \cite{ren2021comprehensive}. 
Similarly, Baker \etal \cite{baker2016designing} and Zhong \etal \cite{zhong2020blockqnn} utilize the Q-learning for NAS. 
In addition to RL, EA is also a practical search strategy, where each neural network in the search space is encoded as a chromosome to be crossed or mutated in the exploration \cite{miikkulainen2019evolving, sun2020automatically}. 
However, these two categories of methods are extremely inefficient in computations, since both the reward in RL and the fitness in EA are obtained by evaluating the architecture performance on validation set, which requires each architecture to be trained from scratch on training set. 
Several works have been proposed to amortize the computing cost of RL and EA, such as weight sharing \cite{pham2018efficient}, knowledge inheritance \cite{zhang2018finding} and proxy metrics \cite{zhou2020econas}. 
The GD-based method is another breakthrough in NAS. 
Liu \etal \cite{liu2018darts} loosen the search space, making the model architecture a continuous variable that can be optimized by efficient gradient descent algorithm. 
Besides, Cai \etal \cite{cai2018proxylessnas} add hardware metrics for searching, Chen \etal \cite{chen2019progressive} propose a progressive way to gradually build search architecture, and Hu \etal \cite{hu2020dsnas} utilize the Gumbel-Softmax to reduce the memory footprint of model searching.

\subsection{\textbf{Deep Long-Tailed Learning}}
In the previous research, a lot of efforts have been made to improve the performance of deep learning on the long-tailed datasets.
These works can be roughly categorized into the following two types \cite{cao2019learning}:

\textbf{Re-Sampling.}
There are two main re-sampling techniques: over-sampling minority classes \cite{byrd2019effect} and under-sampling frequent classes \cite{buda2018systematic}. However, under-sampling has the drawback of discarding a significant portion of data, making it impractical for cases of extreme data imbalance. Conversely, while over-sampling is often effective, it can cause over-fitting of the minority classes. Implementing stronger data augmentation for minority classes can mitigate this over-fitting issue.

\textbf{Re-Weighting.}
Cost-sensitive re-weighting involves assigning adaptive weights to different classes or individual samples. 
A basic approach re-weights classes in proportion to the inverse of their frequency \cite{wang2017learning}. 
However, these methods often complicate the optimization of deep models, especially in scenarios with extreme data imbalance and large-scale datasets. 
Cui \etal \cite{cui2019class} found that re-weighting by inverse class frequency leads to poor performance on frequent classes and thus suggested re-weighting by the inverse effective number of samples. 
Another approach focuses on assigning weights to each sample based on their specific characteristics. 
Li \etal \cite{li2019gradient} proposed a refined method that down-weights samples with either very small or very large gradients, as samples with small gradients are already well-classified and those with large gradients are typically outliers.

\section{Methodology} \label{sec:methodology}

In this work, we mainly focus on searching the convolutional neural networks (CNNs) to solve the computer vision problem (\eg, image classification).
Next, we will introduce some technical details related to the scheme in \cite{timofeev2021self}, including the following three parts:

\begin{figure*}[t]\centering
    \includegraphics[width=0.7\textwidth]{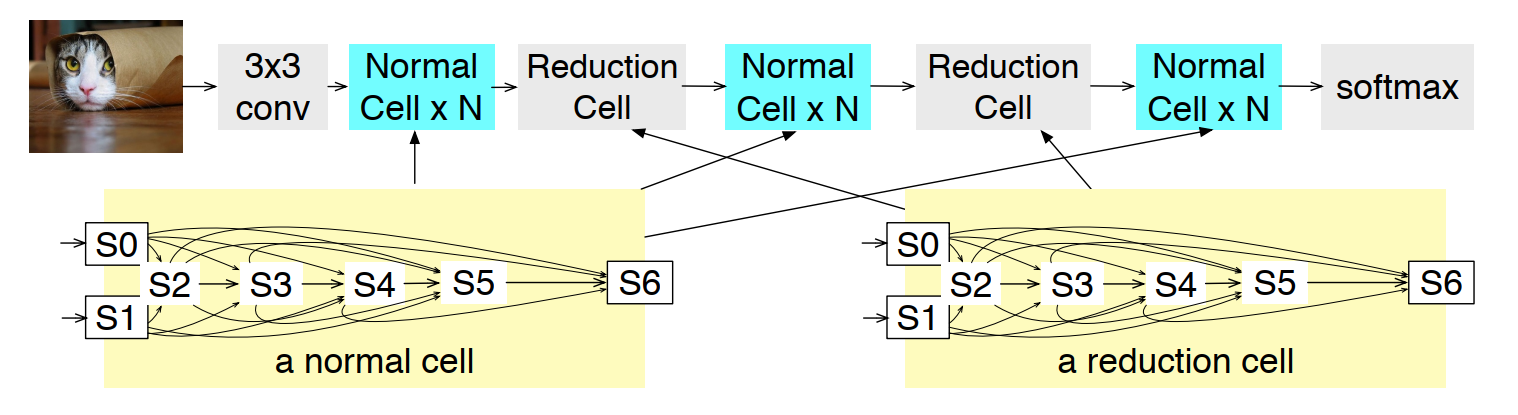}
    \caption{The Illustration of Supernet in DARTS \cite{he2021fednas}.} \label{fig:supernet}
 \end{figure*}

 \begin{figure*}[t]\centering
    \includegraphics[width=0.65\textwidth]{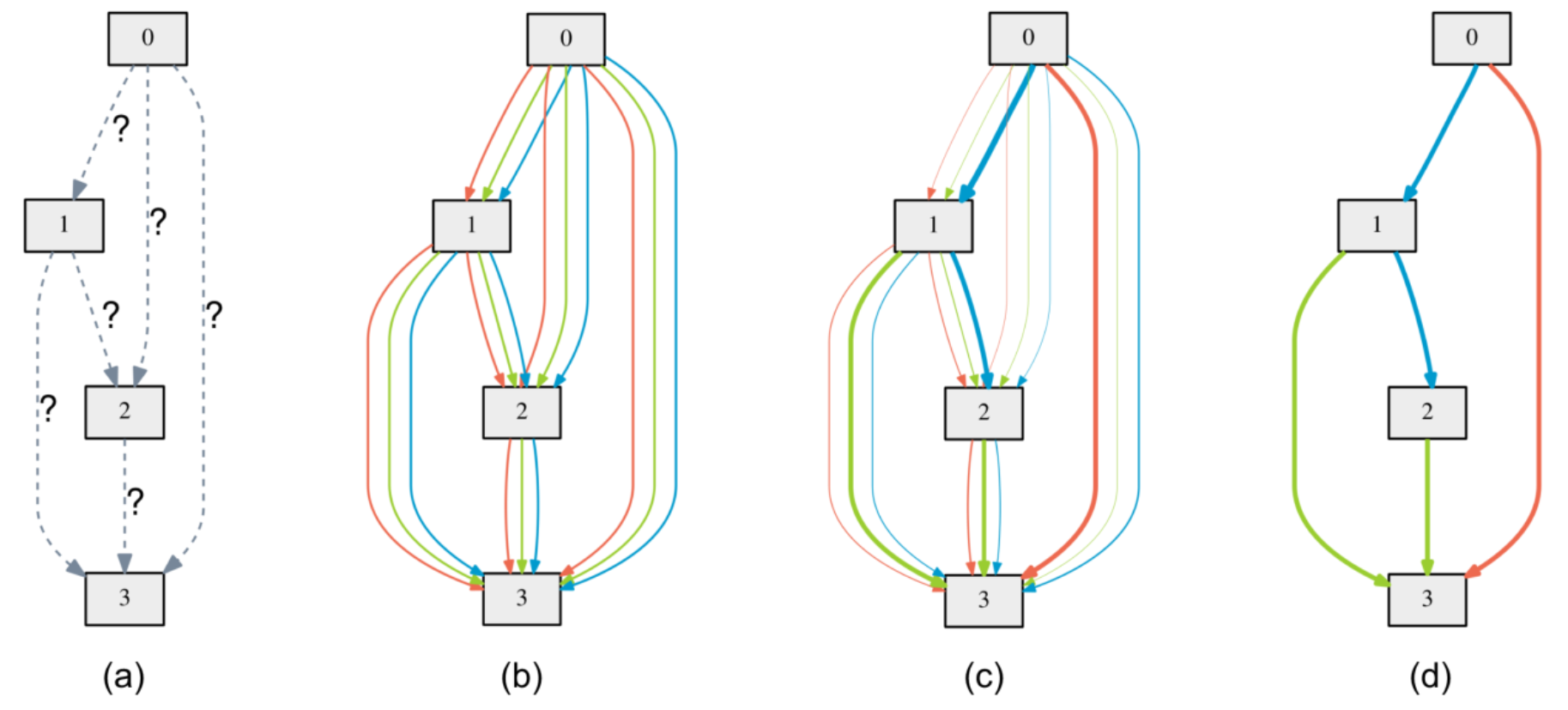}
    \caption{The Process of Searching a Cell's Architecture in DARTS \cite{liu2018darts}.} \label{fig:darts_process}
 \end{figure*}

\subsection{\textbf{DARTS}}
As described in Section \ref{sec:related_work_nas}, both the RL-based and EA-based search strategy would lead to significant computation overhead.
In order to improve the resource efficiency of NAS, Liu \etal \cite{liu2018darts} proposed a differentiable architecture search method, called DARTS, which can accelerate the searching process thousands of times.

The key idea of DARTS is to loose the search space, so that the candidate architectures is no longer discrete, but continuous and differentiable.
As illustrated in Fig. \ref{fig:supernet}, there is a supernet in DARTS, which is stacked using a backbone architecture named cell.
Specifically, cells located at the 1/3 and 2/3 of the total depth of the network are reduction cells to downsample the data and decrease computational complexity.
Other cells are normal cells that perform regular transformations to capture and process feature representations.
The objective of DARTS is to search the architectures of normal cell and reduction cell.
Each cell can be viewed as a directed acyclic graph (DAG), where each node is a feature map, and each directed edge contains several candidate operations to transform
the node.
Take the cell in Fig. \ref{fig:supernet} as an example, a cell contains three types of nodes, which are input nodes (\ie, S0, S1), intermediate nodes (\ie, S2, S3, S4, S5), and output nodes (\ie, S6).
Each cell takes the outputs of the previous two cells as its two input nodes. For each intermediate node in a cell, it connects to all the previous nodes, including the input nodes and other intermediate nodes.
The output node of a cell is the concatenation of all its intermediate nodes.

The process of searching the architecture of a cell is depicted in Fig. \ref{fig:darts_process}, including four steps:

\textbf{(a)} The searching objective is to find the optimal operations on every edge, which are initially unknown.

\textbf{(b)} Let $\mathcal{O}^{(p, q)}$ denote the operation set of edge $(p, q)$. 
Each operation $o$ in $\mathcal{O}^{(p, q)}$ is a function and its input is $x_{p}$ (\ie, the feature map in node $p$).
 In Fig. \ref{fig:darts_process}, assuming that the red line represent convolution operation, the green line represent pooling operations, and the blue line represent residual connection.
Each operation $o$ is associated with an attention weight $\alpha^{(p, q)}_{o}$, which is a differentiable continuous variable and can be trained through gradient descent. 
The edge $(p, q)$ transforms node $p$ to node $q$ by taking softmax over the outputs of the operations in $\mathcal{O}^{(p, q)}$ with the attention weights:
    
\begin{equation} \label{transform_process}
    O^{(p, q)} (x_{p}) = \sum_{o \in \mathcal{O}^{(p, q)}} \frac{ \mathop{\exp} (\alpha^{(p, q)}_{o})}{\sum_{o^{\prime} \in \mathcal{O}^{(p, q)}} \mathop{\exp} (\alpha^{(p, q)}_{o^{\prime}})} o(x_{p})
\end{equation}
where $o(\cdot)$ is the output of operation $o$ and $O^{(p, q)}(\cdot)$ is the value input from node $p$ to node $q$.

\textbf{(c)} In this way, the task of architecture search is reduced to learning a set of continuous variables $\alpha = \{ \alpha^{(i, j)} \}$.
Let $\mathcal{L}_{train}$ and $\mathcal{L}_{val}$ represent the loss value on training dataset and validate dataset, respectively.
The goal for architecture search is to find $\alpha^{*}$ that minimizes the validation loss $\mathcal{L}_{val} (\omega^{*}, \alpha^{*})$, where the weights $\omega^{*}$ associated with the architecture are obtained by minimizing the training loss $\omega^{*} = \arg\min_{\omega} \mathcal{L}_{train}(\omega, \alpha^{*})$.
In conclusion, the architecture search task can be formulated as the following bi-level optimization problem, where $\alpha$ is the upper-level variable and $\omega$ is the lower-level variable:

\begin{align} 
    \min_{\alpha} &\quad \mathcal{L}_{val}(w^*(\alpha), \alpha) \notag \\
    \text{s.t.} &\quad w^*(\alpha) = \arg \min_{w} \mathcal{L}_{train}(w, \alpha) \label{darts_objective}
\end{align}

However, evaluating the architecture gradient exactly can be prohibitive due to the expensive inner optimization.
To this end, DARTS also propose to approximate $w^*(\alpha)$ by adapting $\omega$ using only a single training step, without solving the inner optimization in Eq. \eqref{darts_objective} completely by training until convergence:

\begin{align}
& \nabla_{\alpha} \mathcal{L}_{val}(w^*(\alpha), \alpha) \notag \\
\approx & \nabla_{\alpha} \mathcal{L}_{val}(w - \xi \nabla_{w} \mathcal{L}_{train}(w, \alpha), \alpha)
\end{align}

where $\omega$ denotes the current weights, and $\xi$ is the learning rate for a step of inner optimization.
While training the supernet in this way, the attention weights of the operations that are beneficial to reduce the loss value over the target dataset will be increased.
In Fig. \ref{fig:darts_process}(c), the thickness of the lines indicates the size of the attention weights of the operations, with thicker lines representing larger weights.
Thus, with the supernet training, it can be find that between nodes 0 and 3, the convolution operation (red line) is the optimal choice, while the residual connection (blue line) is the next best option and the pooling operation (green line) is the least effective.

\textbf{(d)} After the supernet achieves convergence, the optimal architecture can be sampled from the DAG of cell according to the attention weights set $\alpha$.
Specifically, in every edge, the operation with the largest attention weight is picked to construct the final architecture:

\begin{equation} \label{eq:softamx}
    o^{(i,j)} = \arg\max_{o \in \mathcal{O}} \alpha_{o}^{(i,j)}
\end{equation}

For example, as shown in Fig. \ref{fig:darts_process}(d), node 0 and node 1 are connected by a residual layer (blue line) in the searched architecture.
It is worth mentioning that it is allowed to choose no operation on an edge.
For instance, node 0 input nothing to node 1.

\subsection{\textbf{FairDARTS}}
Since the supernet involves the weights from all the candidate operations, its training overhead can be very expensive.
Therefore, during the search phase, a shallow supenet is adopted to mitigate the computation cost.
Once the supernet reaching convergence, a deep model is constructed according to the final architecture for the target dataset training.
However, the architecture searched with a shallow supernet may not appropriate for deep
model.
A main reason is that the residual connections are frequently selected as the operation $o^{(i,j)}$.
This is because the residual connection can achieve better performance by leveraging the relationships of other operations, resulting in an unfair advantage of residual connection.
Besides, different operations compete through the softmax normalization function, which further amplifies the unfairness.
To avoid the such unfair competition with residual connection, FairDARTS \cite{chu2020fair} proposed to replace the the softmax normalization with the sigmoid function.
That is to say, the following formula replaces the Eq. \eqref{eq:softamx}:

\begin{equation} \label{eq:sigmoid}
    \bar{o}_{i,j}(x) = \sum_{o \in \mathcal{O}} \sigma(\alpha_{o_{i,j}}) o(x)
\end{equation}

In addition, there is another issue in the search process of DARTS.
After the search process is completed, the operation corresponding to the maximum attention weights is selected and other operations are discarded.
In other word, the continuous attention weights are transformed to the discrete values 0 or 1.
However, such direct transformation can lead to biases in the selection process. 
To this end, FairDARTS used a zero-one loss under fair conditions to push the attention weights toward 0 or 1, thereby reducing the bias present when converting continuous encoding to one-hot encoding.
Let $N$ denote the number of candidate operations in every edge, the proposed zero-one loss can be formulated as follows:

\begin{equation}
    \mathcal{L}_{0-1} = -\frac{1}{N} \sum_{i=1}^{N} \left| \sigma(\alpha_i) - 0.5 \right|.
\end{equation}

With the above zero-one loss, the final loss for the attention weights is:

\begin{equation}
    \mathcal{L}_{total} =  \mathcal{L}_{val}(\omega^*(\alpha), \alpha) + \lambda_{0-1} \mathcal{L}_{0-1}
\end{equation}
where $\lambda_{0-1}$ controls the importance of the zero-one loss.

\begin{figure}[t]\centering
    \includegraphics[width=0.5\textwidth]{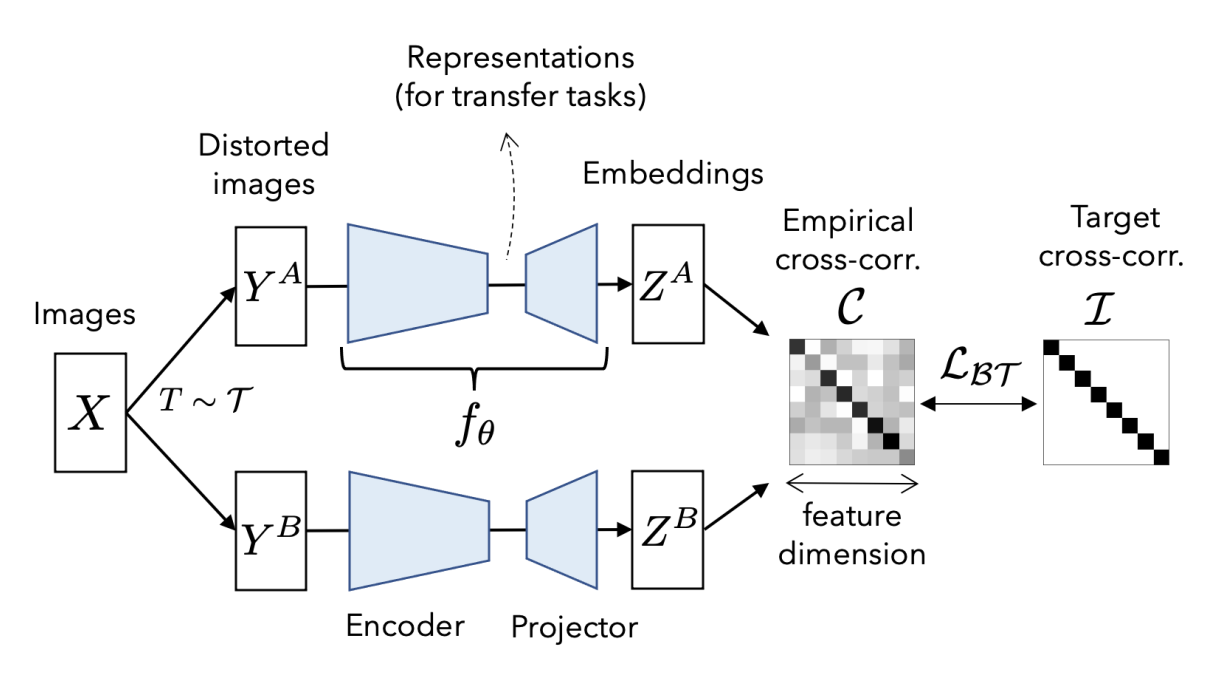}
    \caption{The Depiction of Barlow Twins.} \label{fig:barlow_twins}
 \end{figure}
 
\subsection{\textbf{Barlow Twins}}

As shown in Fig. \ref{fig:barlow_twins}, the Barlow Twins method generates a pair of images from each original image. This pair is formed by applying two different randomly selected transformations (such as random cropping, horizontal flipping, color distortion, etc.) denoted as $Y^{A}$ and $Y^{B}$. This process is repeated for each sample in the mini-batch $X$. The model $f_{\theta}$ then extracts representations $Z^{A}$ and $Z^{B}$ from these distorted versions of the original mini-batch. The objective is to align the cross-correlation between $Z^{A}$ and $Z^{B}$ as closely as possible to the identity matrix. The loss function used for this purpose is defined as:

\begin{equation}
\mathcal{L}_{BT} = \sum_{i} (1 - \mathcal{C}{ii})^2 + \lambda \sum_{i} \sum_{j \ne i} \mathcal{C}_{ij}^2
\end{equation}
where $\lambda$ is a positive scaling factor and $\mathcal{C}$ represents the cross-correlation matrix of the outputs, computed as follows:

\begin{equation}
\mathcal{C}_{ij} = \frac{\sum_b Z_{b,i}^A Z_{b,j}^B}{\sqrt{\sum_b \left( Z_{b,i}^A \right)^2} \sqrt{\sum_b \left( Z_{b,j}^B \right)^2}}
\end{equation}

Here, $b$ indexes the batch samples, and $(i,j)$ indexes the dimensions of the output vectors. Essentially, this method encourages the model to distinguish between the two different images in each pair. The benefit of the Barlow Twins approach is that by decorrelating the outputs, it eliminates redundant information about the samples in the output units.

By the experimental results in \cite{zbontar2021barlow}, Barlow Twins can achieve a high accuracy even in the absence of data labels.
Therefore, it has the potential to enhance the searching and training performance of NAS on the target datasets with long-tailed distributions.

\subsection{\textbf{Integrating FairDARTS and Barlow Twins}}
The authors in \cite{timofeev2021self} propose a novel NAS scheme for the imbalanced datasets (\eg, those with long-tailed label distributions).
For ease of description, we denote this scheme as SSF-NAS.
Firstly, SSF-NAS is build on the top of FairDARTS to avoid unfair competition between residual connections and other candidate operations and enabling DARTS to search better architecture even with a shallow supernet.
Secondly, it has been demonstrate that the self-supervised learning is beneficial for the model training over imbalanced datasets \cite{yang2020rethinking}.
Therefore, the Barlow Twins loss function is adopted to replace the supervised loss in SSF-NAS to improve the searching and training performance on long-tailed datasets.

\section{Experiments} \label{sec:experiments}

\subsection{\textbf{Dataset and Label Distribution}}

In our experiments, we adopt the CIFAR10 dataset \cite{krizhevsky2009learning} for the performance evaluation, including 60,000 32$\times$32 color images evenly from 10 classes, with 50,000 for training and 10,000 for testing.
Due to the limited computation power of our experimental equipment, we were unable to train the medical datasets that is naturally long-tail distributed.
We follow \cite{cao2019learning} to create the long-tailed distribution.
There are three types of label distributions adopted in our experiments:
\begin{itemize}
    \item \textbf{Balance:} Every class contains all the samples from the original dataset, \ie, 5,000.
    \item \textbf{Step:} For the latter half of the classes, the samples number is 5,000 times an imbalance factor $\mu$.
    \item \textbf{Exponential:} For each class $i \in \{0, 1, \cdots, 9\}$, the samples number is 5,000 times $\mu^{i/9}$.
\end{itemize}

\begin{figure}[t]
	\centering

	\subfigure[Balance]{
		\includegraphics[width=0.225\textwidth]{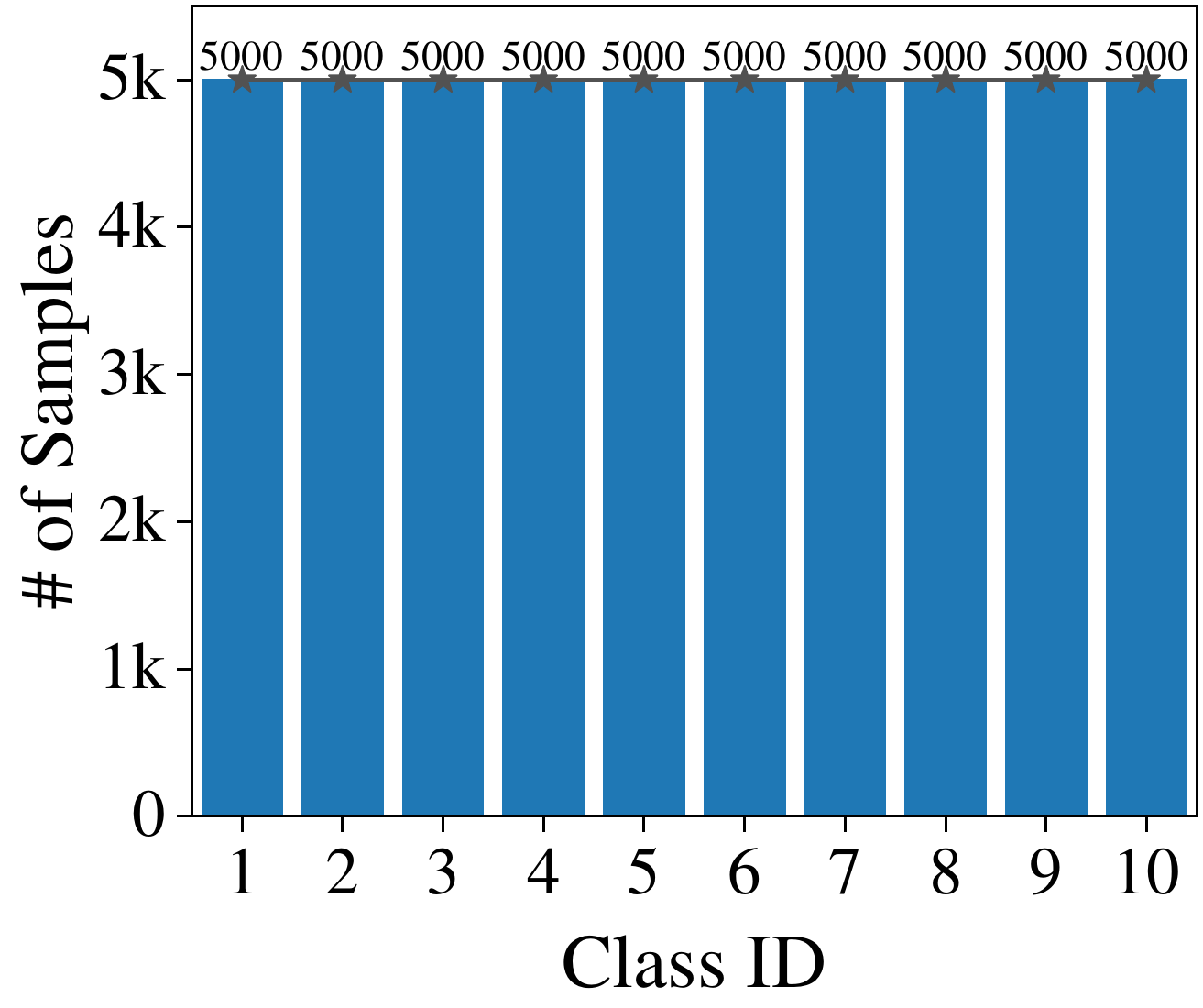}\label{fig:balance_distributon}
	}
	\subfigure[Exponential ($\mu = 0.1$)]{
		\includegraphics[width=0.225\textwidth]{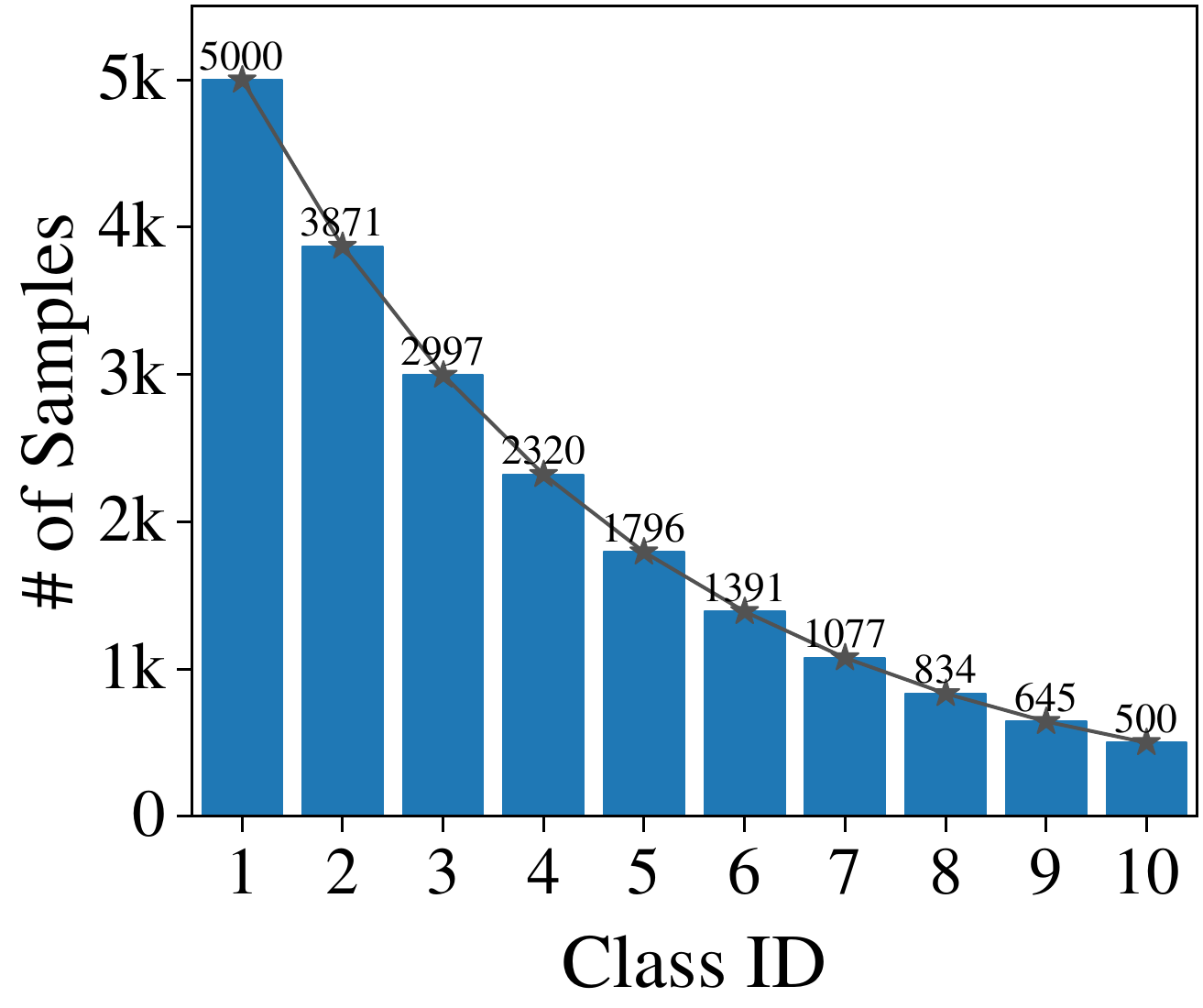}\label{fig:exp01_distributon}
	}
	\subfigure[Exponential ($\mu = 0.01$)]{
		\includegraphics[width=0.225\textwidth]{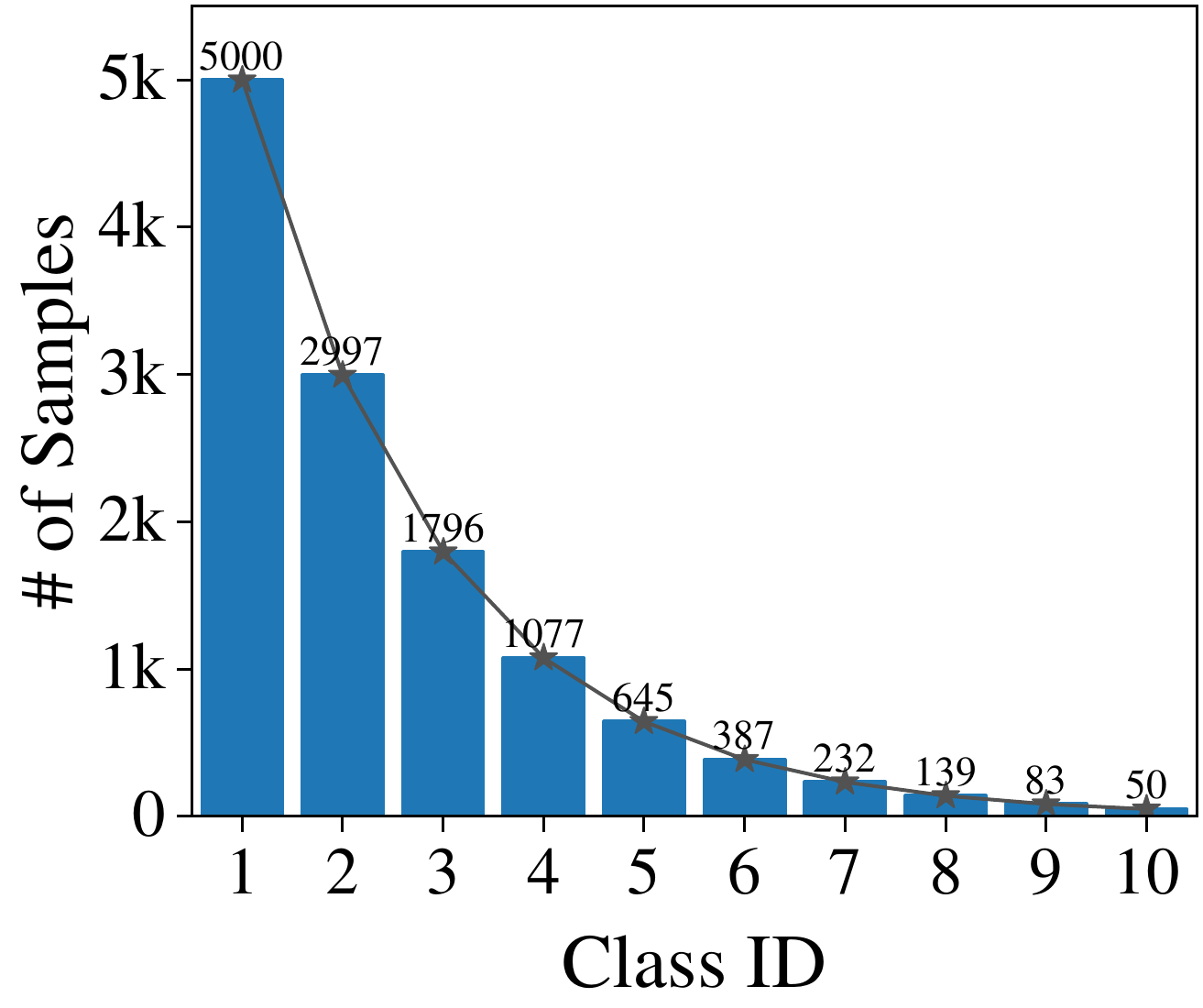}\label{fig:exp001_distributon}
	}
	\subfigure[Step ($\mu = 0.01$)]{
		\includegraphics[width=0.225\textwidth]{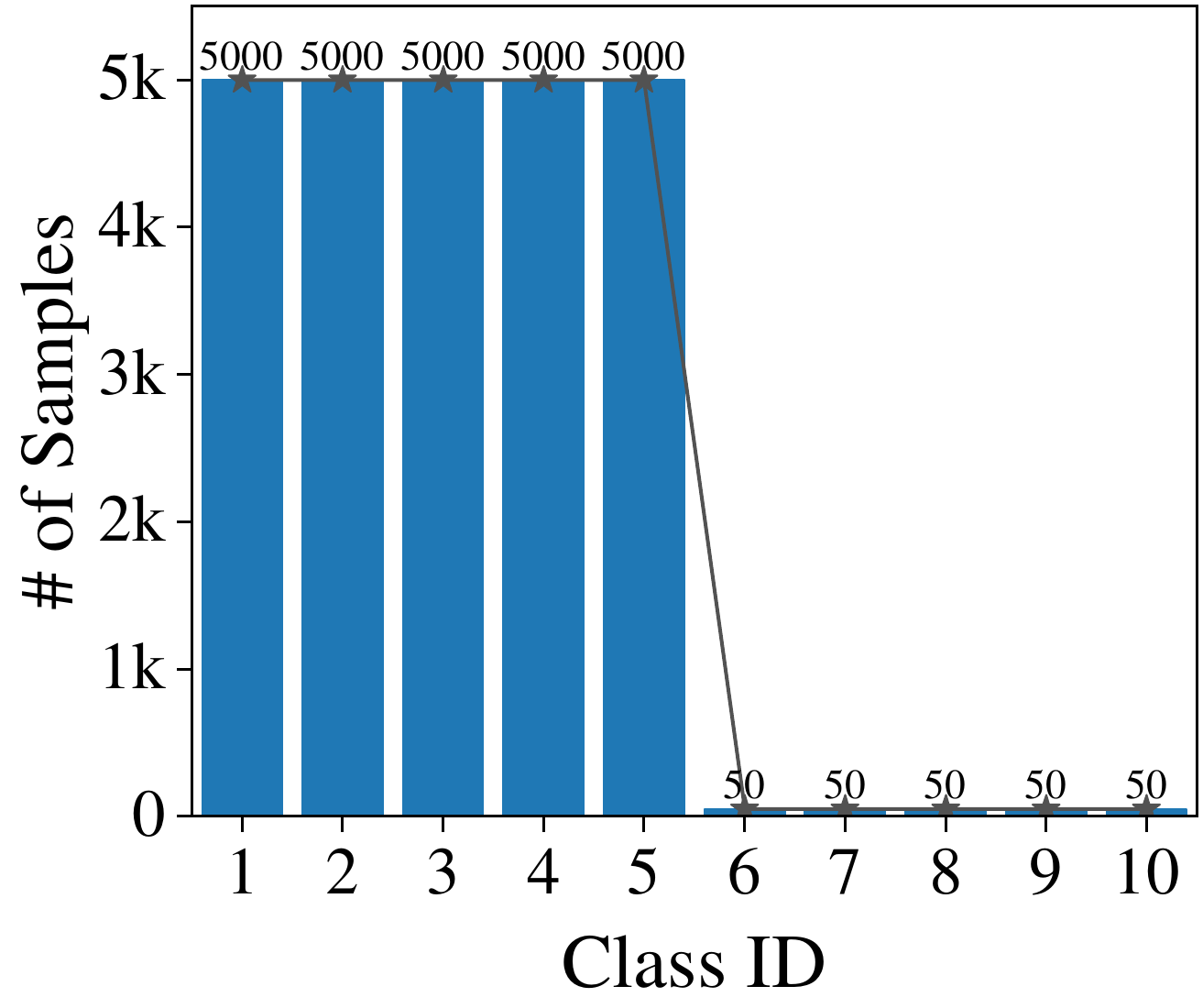}\label{fig:step001_distributon}
	}
	\caption{Four Different Label Distributions in Our Experiments.}\label{fig:label_distribution}
	
\end{figure}

Specifically, our experiments involves four different label distributions, which are illustrated in Fig. \ref{fig:label_distribution}.
By Figs. \ref{fig:exp01_distributon}-\ref{fig:exp001_distributon}, it can be find that the smaller the imbalance factor $\mu$, the more pronounced the long tail characteristics of the label distribution.

\subsection{\textbf{Supernet and Search Space}}
The design of supernet and search space is based on DARTS.
Concretely, the supernet is stacked by 8 cells, including 6 normal cells, and 2 reduction cells.
Each cell contains 2 input nodes, 4 intermediate nodes and 1 output node.
There are 8 operations in every edge of the cell (\ie, search space):
\begin{enumerate}
	\item \textbf{none:} Zero out.
	\item \textbf{max\_pool\_3$\times$3:} Max pooling (window size = 3$\times$3).
	\item \textbf{avg\_pool\_3$\times$3:} Average pooling (window size = 3$\times$3).
	\item \textbf{skip\_connect:} residual connection.
	\item \textbf{sep\_conv\_3$\times$3:} Separable convolution layer with a kernel size of 3$\times$3.
	\item \textbf{sep\_conv\_5$\times$5:} Separable convolution layer with a kernel size of 5$\times$5.
	\item \textbf{dil\_conv\_3$\times$3:} Dilated sepatable convolution layer with a kernel size of 3$\times$3.
	\item \textbf{dil\_conv\_5$\times$5:} Dilated sepatable convolution layer with a kernel size of 5$\times$5.
\end{enumerate}

\subsection{\textbf{Experimental Setup}}

Here we provide our detailed experimental setup.
For the model weights update, we set the learning rate as 0.025 and utilize cosine annealing to adjust it across epochs with a minimum value of 0.001.
Besides, the weight decay and momentum is set to 0.9 and 3$\times$10$^{-4}$, respectively.
For the architecture weights update, a learning rate of 3$\times$10$^{-4}$ and a weight decay of 1$\times$10$^{-3}$ is adopted.
Moreover, the batch size is set to 64.
In each experiment, we train the supernet for 40 epochs for architecture search and then retrain the searched architecture for 150 epochs.

\subsection{\textbf{Experimental Results}}

\subsubsection{\textbf{Verification Experiment}}
\begin{figure}[t]
	\centering
	\subfigure[Balance Test Dataset]{
		\includegraphics[width=0.225\textwidth]{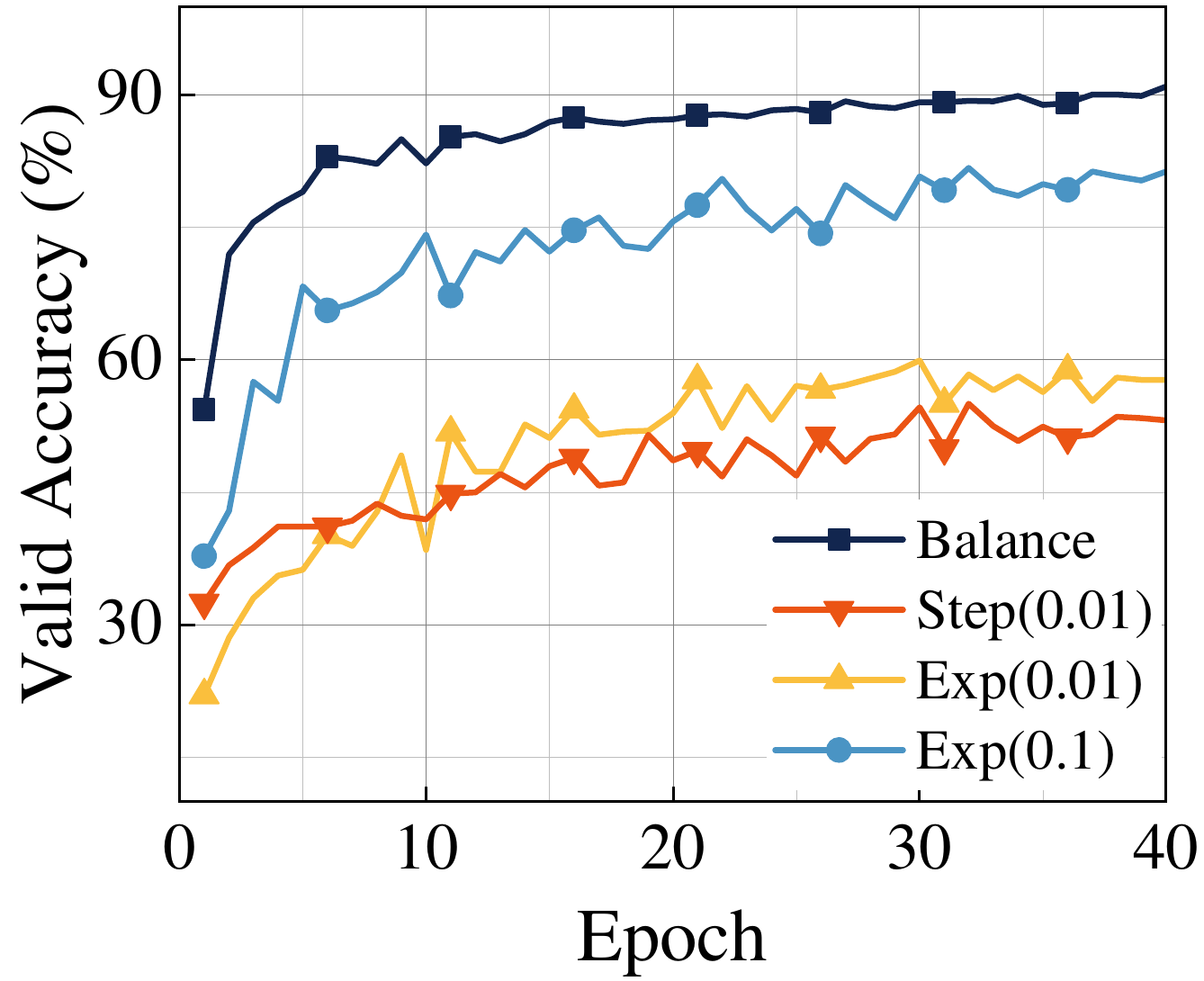}\label{fig:search_b_acc}
	}
	\subfigure[Long-Tailed Test Dataset]{
		\includegraphics[width=0.225\textwidth]{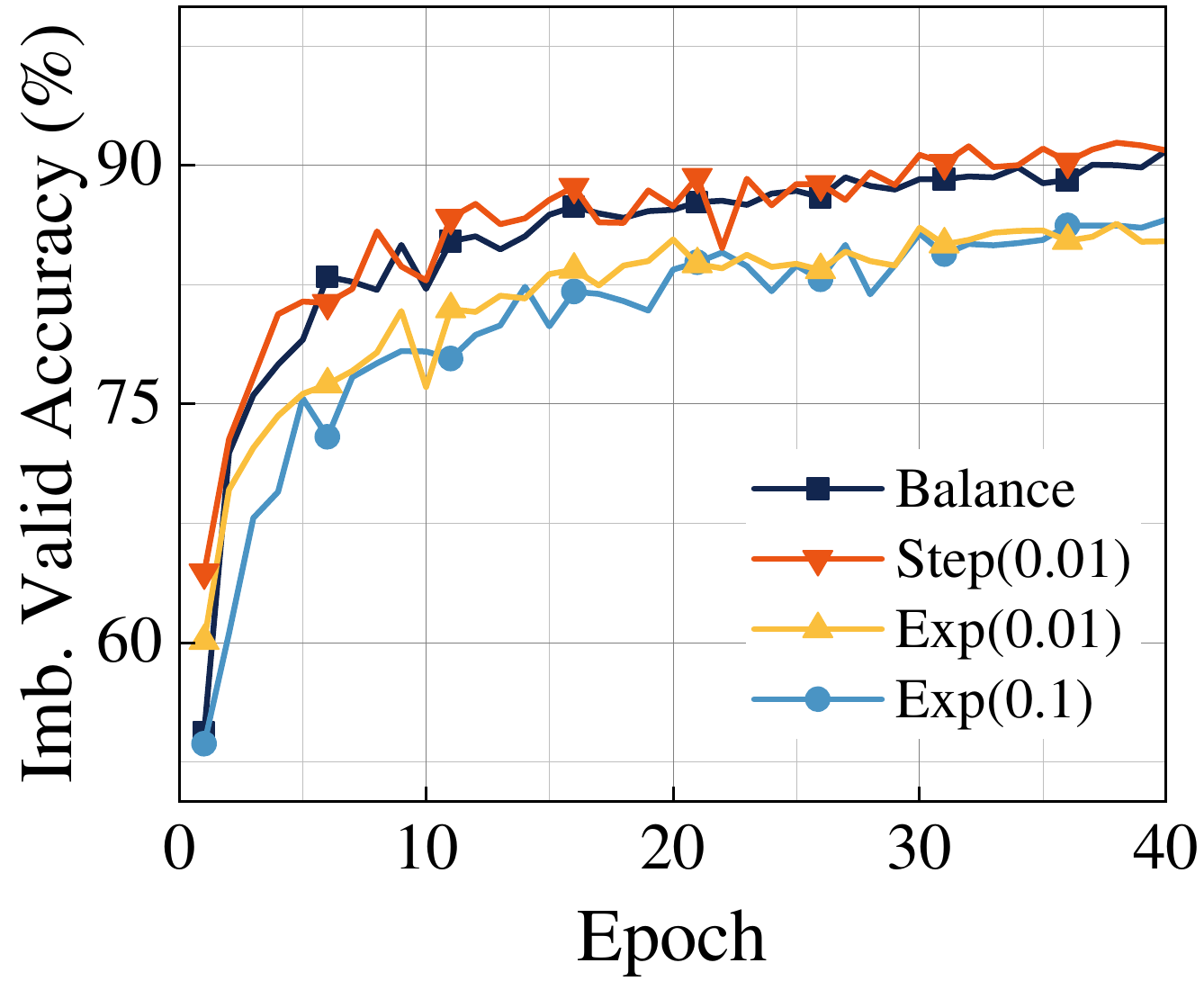}\label{fig:search_imb_acc}
	}
	\caption{Test Accuracy of Supernet in Search Phase.}\label{fig:search_acc}
\end{figure}

\begin{figure}[t]
	\centering
	\subfigure[Test Accuracy vs. Epoch]{
		\includegraphics[width=0.225\textwidth]{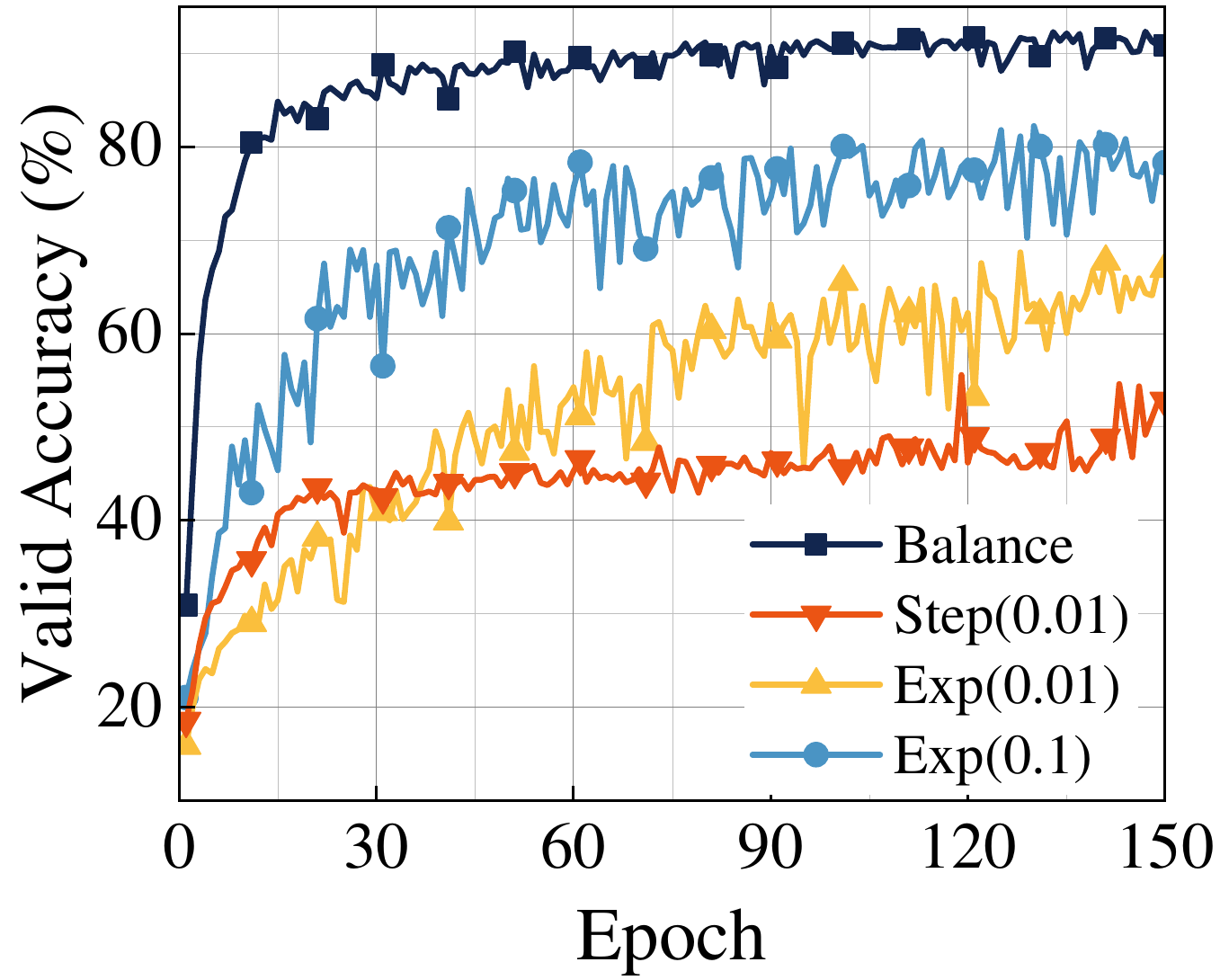}\label{fig:train_acc_1}
	}
	\subfigure[Test Accuracy]{
		\includegraphics[width=0.23\textwidth]{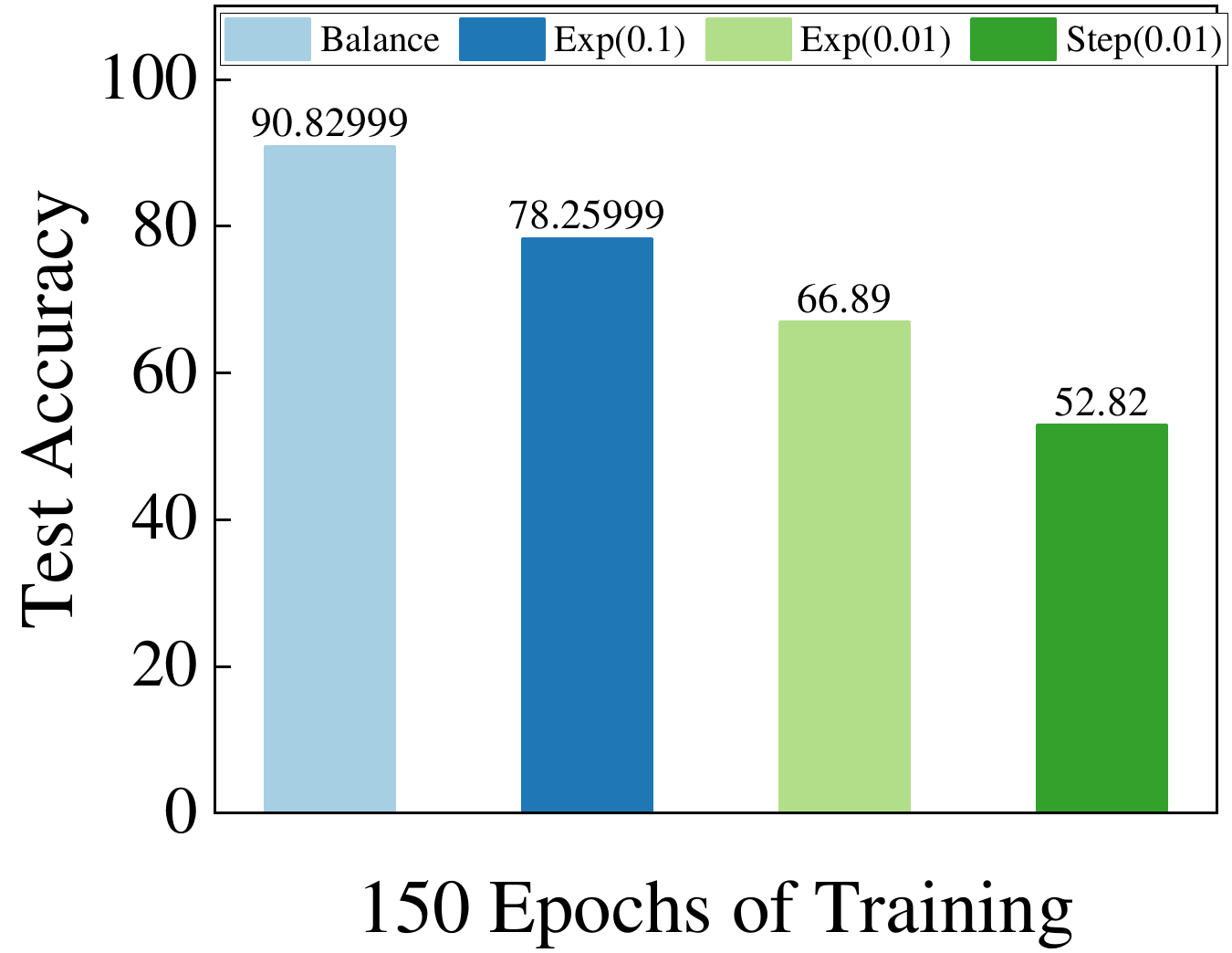}\label{fig:train_acc_2}
	}
	\caption{Test Accuracy of Searched Model in Retrain Phase.}\label{fig:train_acc}
	
\end{figure}

We first conduct a set of experiment to observe the impact of long-tailed distribution on the searching and training performance of NAS.
The results are shown in Figs. \ref{fig:search_acc}-\ref{fig:train_acc}.
From Fig. \ref{fig:search_b_acc}, we can see that the long-tailed data distribution can significantly reduce the searching performance of NAS.
Specifically, while changing the data distribution from balanced to long-tailed, the test accuracy of the supernet shows a significant drop on the balanced test dataset.
However, by Fig. \ref{fig:search_imb_acc}, in the long-tailed test dataset, such accuracy drop is much less than that in the balanced test dataset.
This is because the supernet training is biased towards classes with a lager number of samples, resulting in inadequate fitting for other classes.
The results in Fig. \ref{fig:train_acc} further demonstrate that the architectures searched and trained on long-tailed datasets have a poor generalization performance.
Specifically, as shown in Fig. \ref{fig:train_acc_2}, given the balanced test dataset, the long-tailed distribution will downgrade the valid accuracy by about 12.57\% to 38.01\%.

\subsubsection{\textbf{SSF-NAS Evaluation}}
\begin{figure}[h]
	\centering
	\subfigure[Normal Cell]{
		\includegraphics[width=0.5\textwidth]{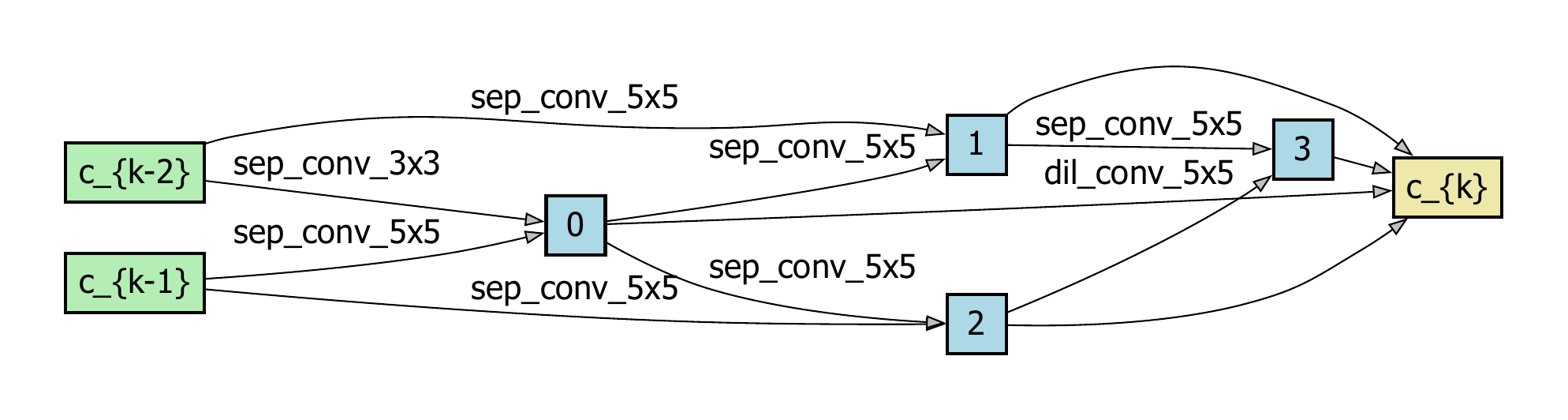}\label{fig:normal}
	}
	\subfigure[Reduction Cell]{
		\includegraphics[width=0.5\textwidth]{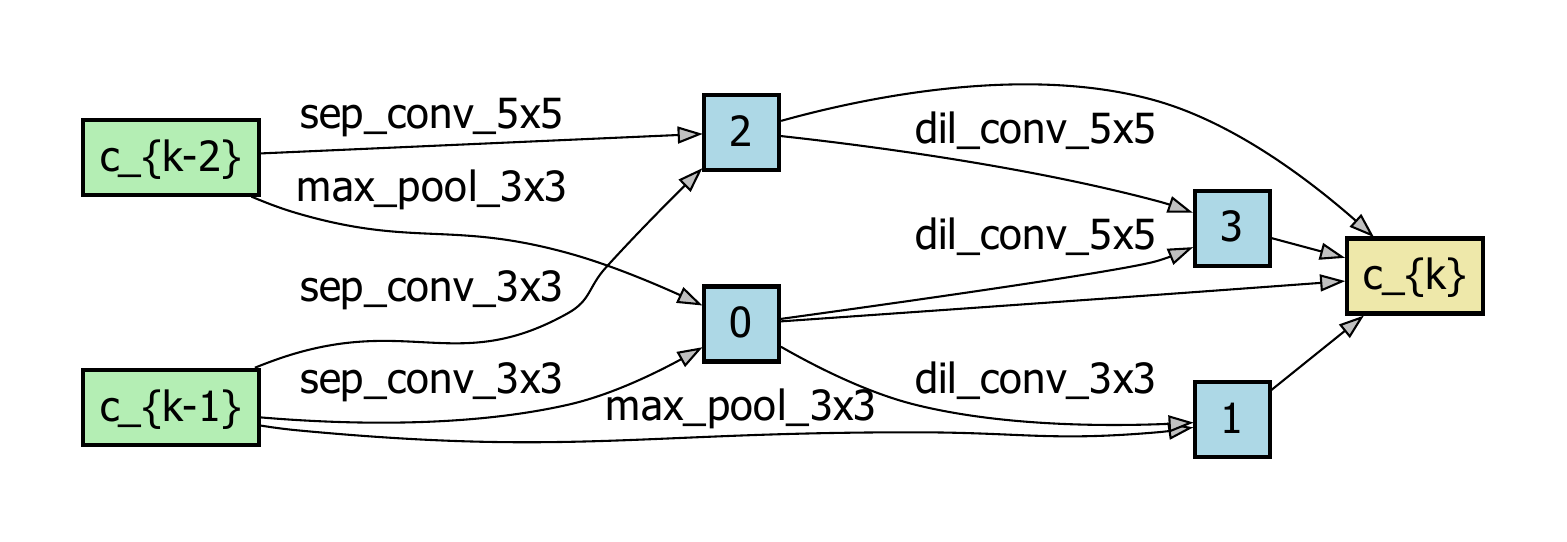}\label{fig:reduction}
	}
	\caption{The Searched Cell Architectures.}\label{fig:cell}
\end{figure}

Further, we replicate the SSF-NAS scheme in our experimental environment, and conduct a set of experiments with the exponential long-tailed distribution and the imbalance factor $\mu$ = 0.1.
The results are shown in Table \ref{tab:ssf-nas}.
Specifically, the architecture searched and trained by SSF-NAS achieve an accuracy of 82.64\% in the balanced test dataset, while DARTS and FairDATRS achieve 78.26\% and 79.35\%, respectively.
Although such improvement is not as significant as the original results in \cite{timofeev2021self}, it is sufficient to demonstrate the effectiveness of SSF-NAS.
The searched architectures of normal cell and reduction cell are shown in Fig. \ref{fig:cell}.
By Fig. \ref{fig:normal}, all the picked operations in the normal cells are convolution layer, which helps to enhance the representation capacity of the model.
Besides, Fig. \ref{fig:reduction} indicates that except for convolution layers, there are also two max pooling operation in the searched reduction cell.
Thus the reduction cell can reduce the dimension of intermediate results and diminish the computation overhead of the subsequent cells.
These experimental results are generally in line with our expectations.

\begin{table}[t]
    \centering
    \caption{Final Test Accuracy of Different Schemes.}
    \begin{tabular}{c|c}
        \toprule
        Schemes & Test Accuracy (\%) \\
        \midrule
        DARTS & 78.26 \\
        \midrule
        FairDARTS & 79.35 \\
        \midrule
        SSF-NAS & 82.64 \\
        \bottomrule
    \end{tabular}
    
    \label{tab:ssf-nas}
\end{table}

\section{Conclusion} \label{sec:conclusion}

In this paper, we aim to improve the searching and training performance of NAS on long-tailed datasets.
We reviewed related works on NAS and deep learning methods tailored for long-tailed datasets, providing a comprehensive understanding of the current landscape.
For reach our target, we do the research on an existing work SSF-NAS, a novel approach that combines self-supervised learning with fair differentiable NAS. 
In order to know SSF-NAS better, we detailed the fundamental techniques underlying SSF-NAS, including DARTS, FairDARTS, and Barlow Twins.
Finally, we conducted a series of experiments on the CIFAR10-LT dataset to verify the impact of long-tailed distribution on NAS, and evaluate the performance of SSF-NAS.
Our results are generally in line with the expectations.

\bibliographystyle{IEEEtran}
\bibliography{main} 

\end{document}